\pdfoutput=1

\documentclass[11pt]{article}

\usepackage[]{ACL}

\usepackage{times}
\usepackage{latexsym}
\usepackage{amsmath}
\usepackage[T1]{fontenc}

\usepackage[utf8]{inputenc}

\usepackage{microtype}

\usepackage{inconsolata}
\usepackage{makecell}

\usepackage{graphicx}
\usepackage{float}
\usepackage{booktabs}
\usepackage{multirow} 
\usepackage{fancyhdr}
\usepackage{listings}
\usepackage{longtable} 
\usepackage{float}
\usepackage{caption}
\usepackage{bm}
\usepackage{hyperref}       
\usepackage{url}            
\usepackage{booktabs}       
\usepackage{array}
\usepackage{xcolor}
\usepackage{amsfonts}       
\usepackage{nicefrac}       
\usepackage{microtype}      
\usepackage{lipsum}
\usepackage{graphicx}       
\graphicspath{{media/}}     
\usepackage{titlesec}
\usepackage[switch]{lineno} 
\usepackage{subcaption}
\usepackage{algpseudocode}
\usepackage{amsmath}
\usepackage{pifont}
\usepackage{svg}
\usepackage[ruled,vlined]{algorithm2e}
\titleformat{\section}{\large\bfseries}{\thesection}{1em}{}

\usepackage{amssymb}
\usepackage[T1]{fontenc}
\newcommand{\cmark}{\textcolor{green!80!black}{\ding{51}}}
\newcommand{\xmark}{\textcolor{red}{\ding{55}}}
\lstdefinestyle{jsonStyle}{
    language=JSON,
    basicstyle=\ttfamily\small,
    numbers=left,
    numberstyle=\tiny,
    stepnumber=1,
    numbersep=8pt,
    showstringspaces=false,
    breaklines=true,
    frame=single,
    keywordstyle=\color{blue},
    stringstyle=\color{green!50!black},
    commentstyle=\color{gray},
    backgroundcolor=\color{gray!10},
    tabsize=2,
    breakatwhitespace=false
}

\title{{C}ond{A}mbig{QA}: A Benchmark and Dataset for Conditional Ambiguous Question Answering}

  \author{
    Zongxi Li\textsuperscript{\rm 1}$^\dagger$\thanks{\ \ Corresponding author; $^\dagger$ \;Equal contribution.}, Yang Li\textsuperscript{\rm 2}$^\dagger$, Haoran Xie\textsuperscript{\rm 1}, S. Joe Qin\textsuperscript{\rm 1} \\
  \textsuperscript{\rm 1} School of Data Science, Lingnan University, Hong Kong SAR \\
  \textsuperscript{\rm 2} School of Science and Technology, Hong Kong Metropolitan University, Hong Kong SAR \\
  \texttt{\{zongxili, hrxie, joeqin\}@LN.edu.hk}\\
  \texttt{\{liya\}@hkmu.edu.hk}\\
 }

\begin{document}
\maketitle
\begin{abstract}

Users often assume that large language models (LLMs) share their cognitive alignment of context and intent, leading them to omit critical information in question-answering (QA) and produce ambiguous queries. Responses based on misaligned assumptions may be perceived as hallucinations. Therefore, identifying possible implicit assumptions is crucial in QA. To address this fundamental challenge, we propose \textbf{Cond}itional \textbf{Ambig}uous \textbf{Q}uestion-\textbf{A}nswering (CondAmbigQA), a benchmark comprising 2,000 ambiguous queries and condition-aware evaluation metrics\footnote{The dataset is available at \url{https://huggingface.co/datasets/Apocalypse-AGI-DAO/CondAmbigQA-2K}.}. Our study pioneers ``conditions'' as explicit contextual constraints that resolve ambiguities in QA tasks through retrieval-based annotation, where retrieved Wikipedia fragments help identify possible interpretations for a given query and annotate answers accordingly. Experiments demonstrate that models considering conditions before answering improve answer accuracy by 11.75\%, with an additional 7.15\% gain when conditions are explicitly provided. These results highlight that apparent hallucinations may stem from inherent query ambiguity rather than model failure, and demonstrate the effectiveness of condition reasoning in QA, providing researchers with tools for rigorous evaluation.
\end{abstract}

\section{Introduction}

Large language models (LLMs) have made remarkable progress in question answering (QA). However, these advanced models remain prone to generate unreliable responses, especially in ambiguous contexts, with hallucinations being a primary concern \cite{ji-etal-2023-towards}. Expectation mismatch is one of several important causes, and its role is especially pronounced when queries omit implicit assumptions and LLMs misinterpret queries due to their limited ability to infer a human-like context \cite{banerjee2024llms}.

Ambiguity in QA is particularly problematic as human communication relies highly on shared background knowledge and implicit cognitive frameworks, often omitting mutual contexts that are not universally recognised outside specific environments. In addition, language itself is inherently ambiguous, as people prefer concise expressions over exhaustive ones \cite{wasow2005puzzle}. 
For example, the seemingly straightforward question ``\textit{When did the US leave the gold standard?}'' admits multiple valid interpretations: a model might answer \textbf{1933}, referring to the suspension of domestic convertibility during the Great Depression; \textbf{1968}, when the legal requirement for gold reserves behind US currency was removed; or \textbf{1971}, when President Nixon ended international convertibility, effectively severing the system globally. Each answer is historically correct but grounded in a different assumption about what it means to ``\textbf{leave}.'' This illustrates how users typically approach QA systems with assumptions that shape intent but remain unstated. Since models lack direct access to these assumptions, responses may be logically sound with the query’s literal wording yet misaligned with user expectations. 
To bridge this gap, we approximate these assumptions by leveraging retrieval to surface possible interpretations, which are formalised as explicit conditions.

We consider that identifying and addressing these implicit assumptions is key to disambiguation, ensuring that generated responses are accurate and aligned with user expectations. Current research focuses on improving model reasoning, expanding context length, and enhancing retrieval and the use of relevant information \cite{shaier-etal-2023-stochastic,ding2024longrope, sun2024determlr}. Techniques such as Chain-of-Thought (CoT) prompting, reinforcement learning \cite{wei2022chain, ahmadian2024back}, and human preference alignment \cite{ji2024beavertails} enhance model capabilities, yet they do not explicitly resolve ambiguity.

This paper introduces \textbf{Cond}itional \textbf{Ambig}uous \textbf{Q}uestion-\textbf{A}nswering (CondAmbigQA), a novel framework that tackles ambiguity by incorporating explicit conditions. To approximate the implicit assumptions underlying ambiguous queries, we use a retrieval-based strategy to surface diverse contextual constraints from external knowledge sources (e.g., Wikipedia). These constraints, defined as ``conditions,'' represent contextual prerequisites that clarify plausible interpretations and pinpoint the answer. Unlike existing datasets that attempt to enumerate all possible answers based on human knowledge, our framework focuses on identifying key conditions that distinguish a question from similar ones. We design a human-LLM interactive annotation process where GPT-4o assists in refining condition-answer pairs, significantly reducing annotation cost and minimising subjectivity.

Using CondAmbigQA, we develop an experimental protocol to evaluate models on both condition identification and conditional answer generation. Our results demonstrate that incorporating explicit conditions into answer generation improves response quality compared to standard retrieval-augmented generation (RAG) methods \cite{lewis2020retrieval}. Larger proprietary models, such as GPT-4o and GLM4-Plus, outperform smaller models in both condition adherence and answer quality. Additionally, we introduce a metric for citation generation, further enhancing answer reliability. Our main contributions are as follows:
\begin{itemize}
    \item We are the first to identify implicit conditions as the root cause of ambiguity in QA tasks and propose a framework for disambiguation through explicit condition representation.
    \item We propose CondAmbigQA, a novel framework that structures QA responses around identified conditions, ensuring clarity and relevance in context-specific answers.
    \item We adopt a human-LLM interactive annotation process that uses GPT-4o to assist in generating condition-answer pairs, significantly reducing annotation costs and maintaining high data quality.
    \item Our experiments highlight the importance of condition in QA, which enables models to achieve substantial improvements in the accuracy of answer generation.
\end{itemize}

\begin{figure*}[ht]
    \centering
    \includegraphics[width=\textwidth, trim={0 1.cm 0 1cm}]{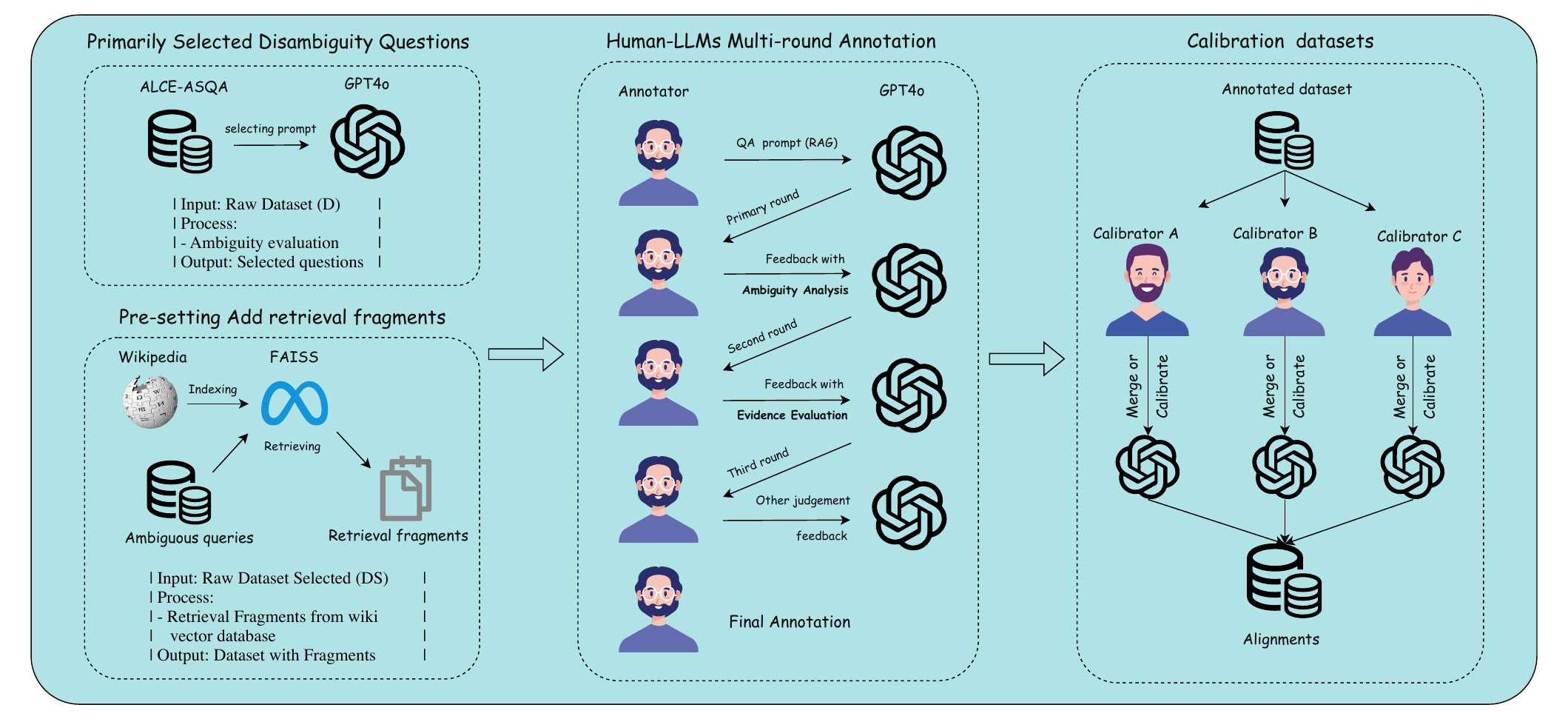}
    \caption{Annotation workflow adopted in CondAmbigQA dataset construction.}
    \label{fig:annotation_process}
\end{figure*}

\section{Related Work}
\label{sec:related_work}

Recent advances in LLM alignment for QA have emphasised interpretability and efficiency through Chain-of-Draft (CoD) prompting \cite{xu2025cod}, reducing verbosity compared to traditional CoT methods. In addition, Process-Supervised Policy Optimisation (PSPO) introduces non-linear reward shaping to balance correctness and brevity in reasoning steps \cite{xu2025cod, li2024pspo}. However, these alignment strategies may embed human-biased rewards, prioritising expected outcomes over proper reasoning \cite{hewitt2024instruction}. 

RAG-based methods have shown promise in improving factual accuracy through retrieval \cite{lewis2020retrieval,gao2023retrieval,li-etal-2025-ese}, 
but they do not directly address ambiguity arising from implicit assumptions. Recently, \citet{zhou2025credible} study the credibility of retrieval-augmented answers in multi-hop scenarios, providing new methods for assessing and improving factual robustness through iterative retrieval strategies. While Self-RAG \cite{asai2024selfrag} and CRAG \cite{yan2024corrective} enhance reliability through reflection or evaluators, newer approaches further refine retrieval credibility, addressing critical gaps in handling complex queries. \citet{liu-etal-2025-open} further points out that conflicting information from retrieved contexts may result in untruthful and inaccurate answers. 

Evaluation of LLM responses presents unique challenges, as traditional metrics like ROUGE and BLEU fail to capture the complexity and nuance of modern model outputs. Several frameworks such as G-Eval \cite{wei2022chain}, self-evolving benchmarks \cite{wang2024benchmark}, LiveBench \cite{white2024livebench}, and MixEval \cite{ni2024mixeval} have emerged. Particularly, \citet{murugadoss2025evaluating} verify the adherence of LLM-based evaluators to task evaluation instructions, offering methodological guidance for robust and precise evaluation. Nevertheless, establishing unbiased and comprehensive metrics remains an ongoing challenge \cite{magesh2024hallucination}.

While existing research has made important advances in ambiguous QA, it faces critical limitations. AmbigQA \cite{min-etal-2020-ambigqa} rewrite ambiguous questions to capture possible answers; however, its reliance on human annotators introduce bias and fails to codify the implicit conditions driving various interpretations. ASQA \cite{stelmakh-etal-2022-asqa} extend AmbigQA by generating long-form answers to cover multiple answers, but its annotation process leads to logical inconsistencies when linking different answer components. ALCE \cite{gao-etal-2023-enabling} enhance credibility through Wikipedia citations, but fail to address the implicit ambiguity within queries. Recent approaches like APA \cite{kim2024aligning} adopt agent-based approaches to prompt users for clarification, but model's internal biases may inadvertently guide users toward unintended choices. BeaverTails \cite{ji2024beavertails} leverage human preference, but this approach can amplify annotation biases. \citet{shaier-etal-2024-adaptive} propose Adaptive Question Answering and identify that ambiguity can be a result of both context ambiguity and question ambiguity.

Unlike prior works that either rewrites queries (AmbigQA, ASQA) or detects ambiguity \textit{post hoc} (APA), our method systematically identifies implicit assumptions by structuring responses around explicit conditions. This approach ensures that retrieved contexts serve as an interpretative guide in reasoning. Furthermore, our condition-aware evaluation provides a more precise evaluation for ambiguity resolution.

\begin{table*}[ht]
\small
\centering
\begin{tabular}{l|c|c|c|c}
\toprule
Dataset & \begin{tabular}[c]{@{}c@{}}Retrieval\\ Included\end{tabular} & \begin{tabular}[c]{@{}c@{}}Complete\\ Answer\end{tabular} & \begin{tabular}[c]{@{}c@{}}Advanced\\ Reasoning\end{tabular} & \begin{tabular}[c]{@{}c@{}}Ambiguity\\ Resolution\end{tabular} \\
\midrule
CondAmbigQA & \cmark & \cmark & \cmark & \cmark \\
\midrule
ASQA \cite{stelmakh-etal-2022-asqa} & \xmark & \cmark & \cmark & \cmark \\
AmbigNQ \cite{min-etal-2020-ambigqa} & \xmark & \xmark & \xmark & \cmark \\
ALCE \cite{gao-etal-2023-enabling} & \cmark & \cmark & \xmark & \xmark \\
Multihop-RAG \cite{tang2024multihoprag} & \cmark & \xmark & \cmark & \xmark \\
NaturalQuestions \cite{kwiatkowski-etal-2019-natural} & \cmark & \xmark & \xmark & \xmark \\
TriviaQA \cite{joshi-etal-2017-triviaqa} & \xmark & \xmark & \xmark & \xmark \\
ELI5 \cite{fan-etal-2019-eli5} & \cmark & \cmark & \cmark & \xmark \\
TruthfulQA \cite{lin-etal-2022-truthfulqa} & \xmark & \cmark & \cmark & \xmark \\
\bottomrule
\end{tabular}
\caption{Comparison of CondAmbigQA with other datasets. }
\label{tab:mcaqa-comparison}
\end{table*}

\section{Dataset Construction and Overview}

\subsection{Definition of ``Condition''}
\label{sec:def_cond}
We first formally define \textbf{conditions} as \textit{a set of contextual constraints that must be satisfied for an answer to be considered correct within a particular scope}. Conditions naturally emerge in RAG systems when retrieved documents provide valid grounds for an answer. The need for conditions arises when users pose questions that yield multiple valid answers \cite{qian2024tell} and thus require clarification.
For example, the question ``when did US currency leave the gold standard?'' yields multiple answers due to the progressive transition in monetary policy. Some may cite the 1933 suspension during the Great Depression, others the 1968 repeal of gold reserve requirements, and still others the 1971 Nixon Shock. The conditions clarify why multiple answers exist by explicitly identifying the underlying constraints, allowing users to understand the holistic context rather than focusing on a single date.

\subsection{Dataset Composition and Structure}
The CondAmbigQA dataset consists of 2,000 annotated instances derived from the ALCE-ASQA\footnote{\scriptsize{\url{https://huggingface.co/datasets/princeton-nlp/ALCE-data}}} \cite{gao-etal-2023-enabling}, which originates from AmbigNQ\footnote{\scriptsize{\url{https://huggingface.co/datasets/sewon/ambig_qa}}} \cite{min-etal-2020-ambigqa}. Each instance contains a user query, retrieved document fragments from Wikipedia\footnote{\scriptsize{\url{https://huggingface.co/datasets/wikimedia/wikipedia}\url{https://huggingface.co/datasets/sewon/ambig_qa}}}, and a structured set of condition-answer-citation triples. The components are formally organised as:

\begin{equation*}
\small
\begin{split}
  \texttt{Query} &\vert \{\texttt{RetrievalDocs}\}:\\ \{ &(\texttt{Condition}_1, \texttt{Answer}_1, \{\texttt{Citation}_1^1, \dots\}), \\
  &(\texttt{Condition}_2, \texttt{Answer}_2, \{\texttt{Citation}_2^1, \dots \}), \\
  &\dots \}.
\end{split}
\end{equation*}
This structure represents a significant advancement over existing datasets by incorporating retrieved documents and explicit conditions, enabling a more fine-grained evaluation of ambiguity resolution. An example of annotated data sample is provided in Appendix~\ref{appendix_label}.

\subsection{Annotation Process and Guidelines}
Figure~\ref{fig:annotation_process} depicts our annotation workflow, which integrates human expertise with LLM capabilities to construct a robust dataset. Identifying conditions from retrieval results and consistently summarising key contextual factors is a highly tedious task for human annotators, making the annotation inherently complex and labour intensive. 
To address this challenge, we leverage LLMs' superior text comprehension abilities to streamline annotation while maintaining human oversight. LLMs can efficiently process retrieved contexts and generate initial condition summaries in a consistent manner, significantly reducing the cognitive load on human annotators and minimising subjectivity. However, careful human validation is still needed, particularly when distinguishing subtle variations leading to different answers \cite{geva2019we}.

The annotation team comprises four full-time PhD candidates and two research assistants from local universities, all specialising in NLP. The first phase involves an initial screening to identify genuinely ambiguous questions. By analysing both the questions and their corresponding long-form answers from ASQA (detailed in Appendix~\ref{appendix_labelb}), we employ GPT-4o to filter out cases where ambiguity does not lead to meaningfully different answers, so that human annotators can focus on cases where ambiguity is truly impactful.

We adopt a triple-round annotation process, where GPT-4o and human annotators iteratively refine the annotations. In the first round, GPT-4o processes each query using predefined dataset-construction prompts to draft initial condition-answer pairs. The complete sets of prompts provided to annotators are listed in Appendix~\ref{appendix_labelc}. Annotators then leverage LLMs to analyse these pairs and validate their ambiguity using given prompts. In the final round, the LLM maps these condition-answer pairs to supporting citations from retrieved passages. Human annotators independently review all the responses, focusing on reasoning coherence, logical soundness, and citation accuracy. If additional information or clarification is needed for more precise tuples, the annotators reject the current output and provide feedback for calibration. If no further refinement is required, the tuples are accepted as final. To ensure data quality, regular team meetings are held to collectively discuss difficult cases and maintain consistency across annotators. 

Through this triple-round process, GPT-4o generates satisfactory condition-answer-citation tuples for $40\%$ of cases without modification. With two additional rounds of expert feedback and calibration, this percentage increased to $85\%$ with Cohen's $\kappa$ approximately 0.72, indicating that although LLMs can handle a substantial portion of the task, human expertise remains essential for handling more complex cases. The finding also suggests that this is a meaningful and challenging research problem, suggesting the need for further studies in condition-guided ambiguity resolution.

The dataset of 2,000 instances reflects a significant scaling effort while maintaining quality. Our LLM-assisted approach drastically improved annotation efficiency, with a total labelling cost of approximately \$800 on API (around \$0.3 to \$0.5 per instance) and time of 80 hours for the entire dataset. This represents substantial cost savings compared to fully manual annotation, which requires at least 30 minutes per query and would have been prohibitively expensive at this scale. 

\subsection{Dataset Features and Advantages}
CondAmbigQA provides a framework for assessing ambiguous QA, incorporating key features that enable systematic evaluation, as outlined in Table~\ref{tab:mcaqa-comparison}.

First, \textbf{retrieval-included} annotations ensure that these different models are evaluated under consistent background information. The retrieved fragments provide evidence for answers and serve as sources for extracting conditions, allowing for assessing how well models utilise contextual information to ground their reasoning. 

Second, CondAmbigQA is designed to ensure \textbf{complete answers} by providing explicit condition-answer-citation pairings. 
Unlike datasets that force a single answer, our structure enables the evaluation of multiple interpretations grounded in conditions, ensuring that answers are both comprehensive and contextually appropriate. Our approach also builds on recent advances in source attribution and citation generation \cite{shaier-etal-2024-adaptive}, further enhancing answer reliability.

Third, the dataset requires \textbf{advanced reasoning} by presenting scenarios that demand nuanced condition identification and answer generation. This challenges models to engage in deeper logical reasoning, encouraging them to generate well-grounded responses. 

Finally, CondAmbigQA emphasises \textbf{ambiguity resolution}, explicitly capturing possible clarifications for ambiguous questions. This allows for a structured evaluation of how effectively models recognise, interpret, and resolve ambiguity by interpreting distinct possible meanings. Compared to other datasets like ASQA and AmbigNQ, CondAmbigQA's unique features makes it particularly well-suited for benchmarking models on ambiguous QA.

\subsection*{Data Sources and Licensing}

CondAmbigQA is built upon AmbigNQ \cite{min-etal-2020-ambigqa}, distributed under the CC BY-SA 3.0 license. Context passages from Wikipedia are under the same license, allowing for reproduction and distribution with appropriate attribution. To maintain consistency with these data sources, we will release our dataset under the CC BY-SA 4.0 license.

\section{Experimental Design}

\subsection{Evaluation Metrics}

To quantitatively assess model performance at each stage, we employ a multi-metric evaluation framework. Let ($M$) denote the model output and ($G$) the corresponding ground-truth. We define G-Eval \cite{liu2023g} to measure the quality of output relative to the reference, following criteria similar to those in \citet{yao2024clave,liu2023g}, as implemented in the \texttt{DeepEval} package\footnote{\scriptsize{\url{https://github.com/confident-ai/deepeval}}}. Four metrics are defined, with detailed prompts provided in Appendix~\ref{appendix_labeld}, which describe the instructions used for LLMs to generate relevant outputs. 
Human evaluation on a small subset (detailed in Appendix~\ref{appendix_labele}) indicates strong correlations between G-Eval and human judgement.

\textbf{Condition Score} quantifies the quality of condition identification by comparing the model's extracted conditions against the ground-truth conditions. It assesses both the completeness and clarity of the extracted conditions. The G-Eval framework evaluates whether the model has accurately identified and clearly articulated all relevant conditions.

\textbf{Answer Score} 
evaluates the factual accuracy and contextual relevance of generated answers by comparing the model's answers against the ground-truth answers. The G-Eval framework assesses whether the responses are factually correct and appropriately address the identified conditions.

\textbf{Citation Score} measures source attribution accuracy, which is defined as follows:
\begin{equation}
\resizebox{\columnwidth}{!}{$
\textit{Citation Score}(M,G) = \frac{\left|\{c \in M.\mathrm{citations}\} \cap \{c \in G.\mathrm{citations}\}\right|}{\left|\{c \in M.\mathrm{citations}\}\right|}.
$}
\end{equation}
This recall-focused metric favours models for citation accuracy over exhaustiveness, i.e. how many attributed citations are actually relevant. 

\begin{table*}[ht]
\centering
\resizebox{.8\textwidth}{!}{%
\begin{tabular}{lccccc}
\hline
Model             & \makecell{Condition\\Score}      & \makecell{Answer\\Score}       & \makecell{Citation\\Score}      & Combined         & \makecell{Diff. of\\Ans. Count}   \\
\hline
\multicolumn{6}{l}{\textbf{API Models}} \\
\hline
GPT-4o            & \textbf{\bm{$0.552$} $\pm$ $0.190$} & \textbf{\bm{$0.558$} $\pm$ $0.157$} & \textbf{\bm{$0.875$} $\pm$ $0.207$} & \textbf{\bm{$0.662$}} & $-0.17$  \\
GLM4-plus         & $0.302$ $\pm$ $0.069$            & $0.420$ $\pm$ $0.097$            & $0.441$ $\pm$ $0.261$            & $0.388$             & $+1.01$  \\
\textit{API Average} & $0.427$                          & $0.489$                          & $0.658$                          & $0.525$             & $+0.42$    \\
\hline
\multicolumn{6}{l}{\textbf{Local Models}} \\
\hline
Qwen2.5 (7B)      & $0.235$ $\pm$ $0.120$            & $0.287$ $\pm$ $0.161$            & $0.558$ $\pm$ $0.359$            & $0.360$             & $-0.45$  \\
DeepSeek-R1 (7B)  & $0.245$ $\pm$ $0.112$            & $0.293$ $\pm$ $0.142$            & $0.501$ $\pm$ $0.342$            & $0.346$             & $+0.36$  \\
GLM4 (9B)         & $0.231$ $\pm$ $0.071$            & $0.290$ $\pm$ $0.090$            & $0.320$ $\pm$ $0.215$            & $0.280$             & $+1.08$  \\
LLaMA3.1 (8B)     & $0.232$ $\pm$ $0.076$            & $0.252$ $\pm$ $0.093$            & $0.306$ $\pm$ $0.246$            & $0.264$             & $+0.94$  \\
Mistral (7B)      & $0.196$ $\pm$ $0.060$            & $0.231$ $\pm$ $0.079$            & $0.263$ $\pm$ $0.214$            & $0.230$             & $+1.09$  \\
Gemma2 (9B)       & $0.170$ $\pm$ $0.091$            & $0.203$ $\pm$ $0.118$            & $0.217$ $\pm$ $0.277$            & $0.197$             & $+0.14$  \\
\textit{Local Average} & $0.218$                          & $0.259$                          & $0.361$                          & $0.280$             & $+0.53$    \\
\hline
\end{tabular}%
}
\caption{Main experiment scores, with separate averages for API and local models, highlighting overall model rankings and performance gaps.}
\label{tab:performance-metrics-separated}
\end{table*}

In addition, two metrics are adopted to evaluate the ability to correctly identify multiple ambiguities. \textbf{Answer Count} captures the actual number of generated answers. \textbf{Count Difference} measures how many more or fewer responses a model generates compared to the expected number, with positive values (e.g., GLM4-plus: $+1.01$) indicating overgeneration and negative values (e.g., GPT-4o: $-0.17$) showing undergeneration of responses.

\textbf{Combined Score} provides an overall evaluation by aggregating the Condition Score, Answer Score, and Citation Score into a single metric. It incorporates calibration mechanisms to address discrepancies in the number of condition-answer pairs generated versus the ground-truth. Penalties are applied for overgeneration, undergeneration, and especially for producing only a single answer pair, indicating failure to recognize ambiguity. The final score is computed as a weighted average of the three core metrics, adjusted by these penalties, ensuring a fair comparison across models with varying generation behaviours. This scoring mechanism encourages models to match GPT-4o's ground-truth-consistent behaviour and balances precision and completeness in conditional QA evaluation.

\subsection{Experimental Protocol}

The experiment protocol comprises two settings. In the primary setting, each model is provided with a query \(Q\) along with the retrieved passages \(P\), and is required to (i) extract disambiguating conditions from \(P\), and (ii) generate answers based on the extracted conditions, supported with citations. The outputs are then evaluated using the aforementioned metrics. This end-to-end evaluation assesses the model's ability in both condition identification and conditional answer generation. Additionally, models are provided with ground-truth conditions alongside \(Q\) and \(P\) in an alternative setting. By comparing the performance of the model-generated and ground-truth conditions, we quantitatively assess the impact of explicit condition guidance on answer generation quality and citation accuracy.

\subsection{Baseline Models and Deployment}
We evaluate seven LLMs of varying sizes and capacities on CondAmbigQA benchmark. This includes two proprietary API-based models, i.e. GPT-4o and GLM4-plus, and five locally-deployed open-source models, i.e. LLaMA3.1 (8B)~\cite{dubey2024llama}, Mistral (7B)~\cite{jiang2023mistral}, Gemma (9B)~\cite{team2024gemma}, GLM4 (9B)~\cite{glm2024chatglm}, Deepseek-R1 (7B)~\cite{guo2025deepseek} and Qwen2.5 (7B)~\cite{yang2024qwen2}. The open-source models are deployed via the \texttt{ollama} framework using default sampling parameters and an $8K$ context window. The models are prompted according to the instructions described in Appendix~\ref{appendix_labeld}.

\section{Experimental Results}

\begin{figure}[htbp]
    \centering
    \includegraphics[width=0.49\textwidth, trim={0 1.cm 0 1.cm},clip]{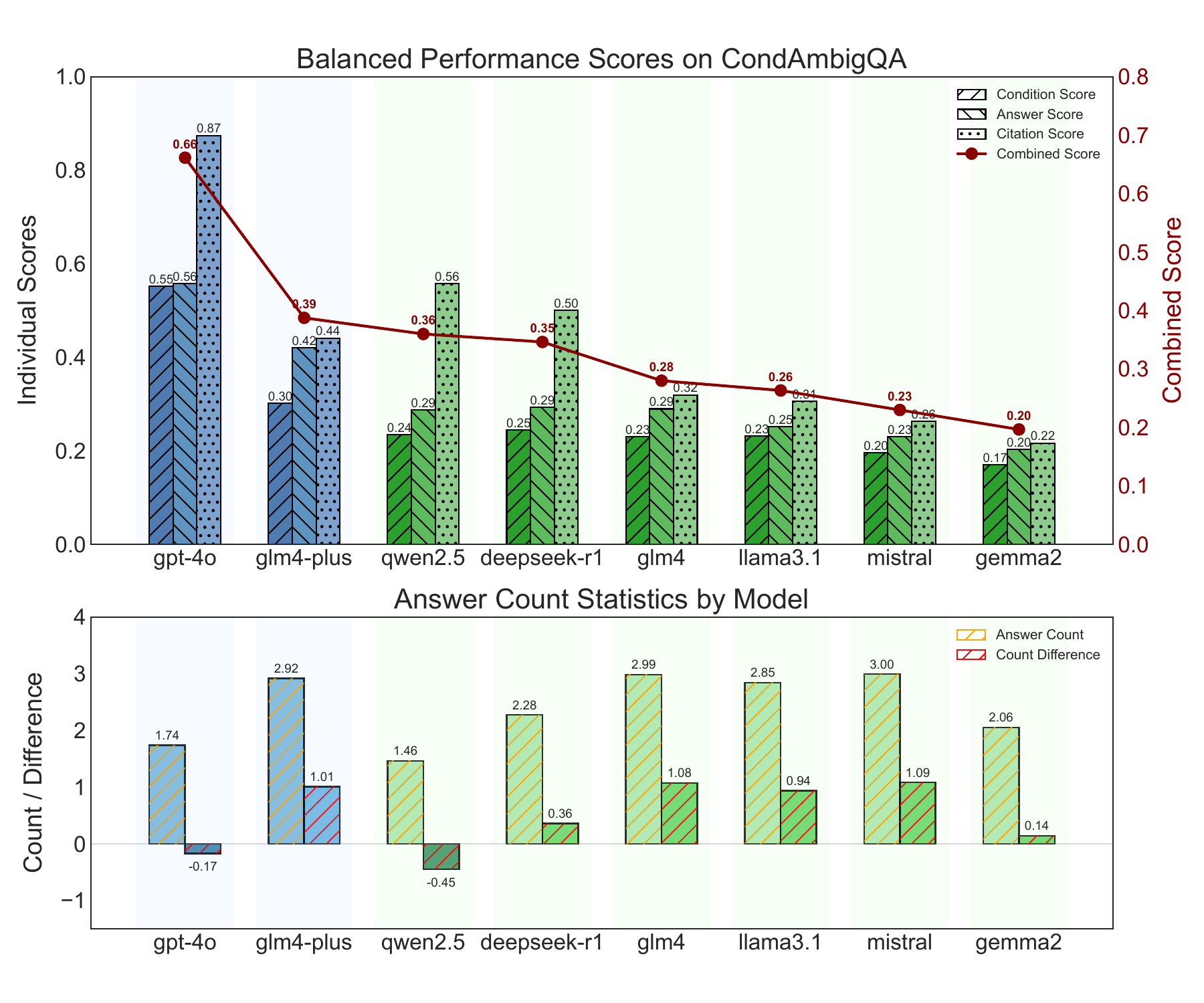}
    \caption{Model performance on four metrics. In particular, it illustrates the relationship between performance and answer count, revealing how different models balance completeness and conciseness.}
    \label{fig:score-bar}
\end{figure}

\subsection{Condition Generation Performance}

The results summarised in Table~\ref{tab:performance-metrics-separated} show significant variability in condition generation capabilities across models. GPT-4o clearly outperforms other models with a condition score of $0.552$ ($\sigma = 0.190$), more than double the average performance of locally-deployed models. Local models showed modest performance, with DeepSeek-R1 at $0.245$, Qwen2.5 at $0.235$, and LLaMA3.1 at $0.232$. Weak performance was observed in Gemma2 at $0.170$ and Mistral at $0.196$. These substantial performance gaps suggest that proprietary API models, particularly GPT-4o, possess enhanced capabilities to identify potential conditions for ambiguous queries, with nearly three times the condition identification capacity of the weakest local models.

We observed that models often struggle to fully capture the context in condition generation. For the query ``when did US currency leave the gold standard?'' (example in Section \ref{sec:def_cond}), Gemma2 generated conditions focusing on ``abandonment of the gold standard in the early 20th century'' ($\text{score} = 0.37$), which captures only the initial phase of the transition without addressing critical later developments. Meanwhile, LLaMA3.1's response emphasised the Great Depression era suspension but failed to articulate the distinction between temporary suspension and final abandonment ($\text{score} = 0.48$). These examples demonstrate that while local models can identify individual historical events, they share common limitations in capturing the bigger picture over time, as reflected in their condition scores rarely exceeding $0.5$.

\begin{figure}[ht]
    \centering
    \includegraphics[width=.95\linewidth, trim={0 1.cm 0 1.5cm}]{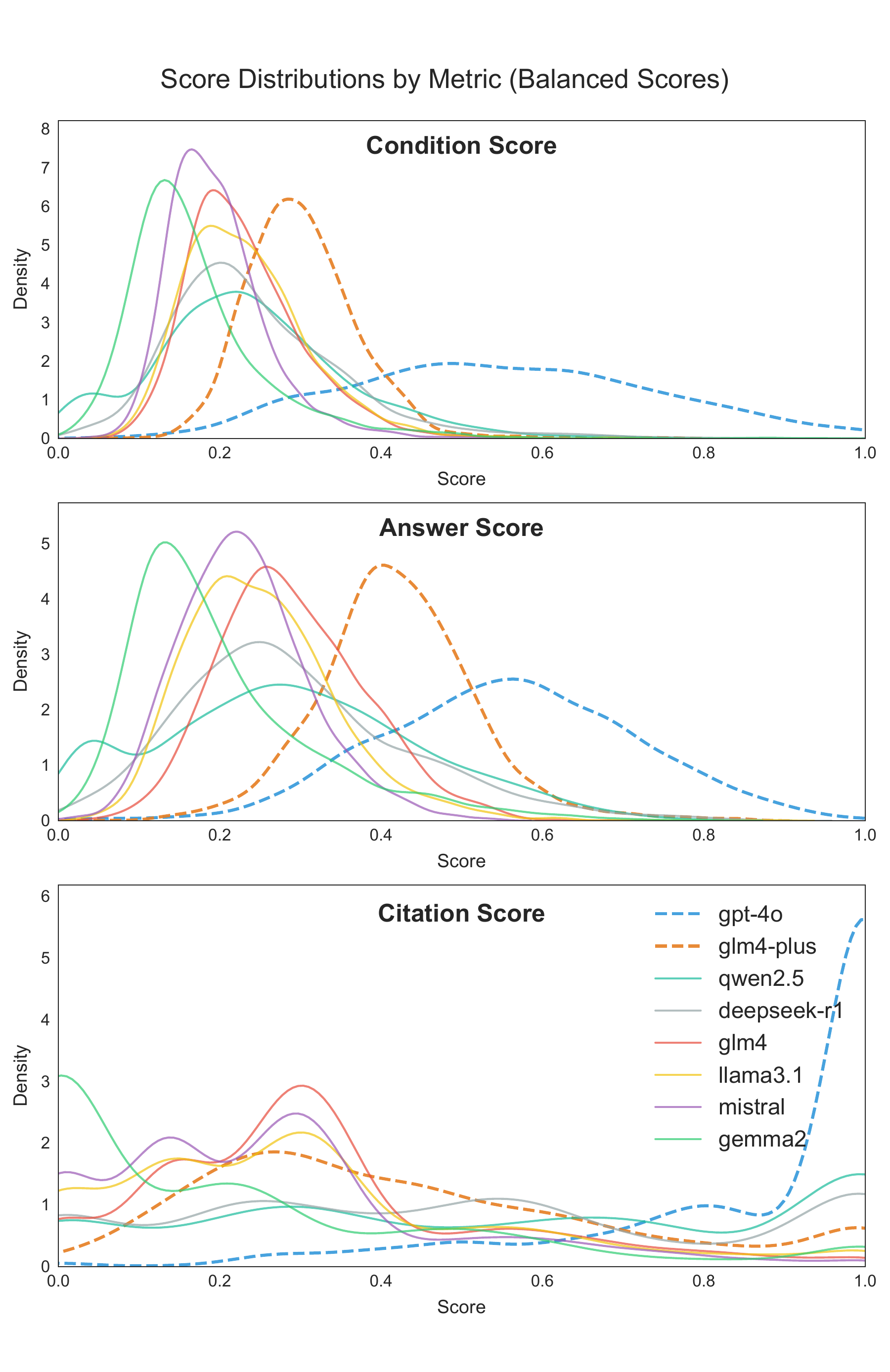}
    \caption{Comparison of score distributions across metrics for models of different scales.}
    \label{fig:expand-distributions}
\end{figure}

\begin{figure*}[t]
    \centering
    \includegraphics[width=.9\textwidth]{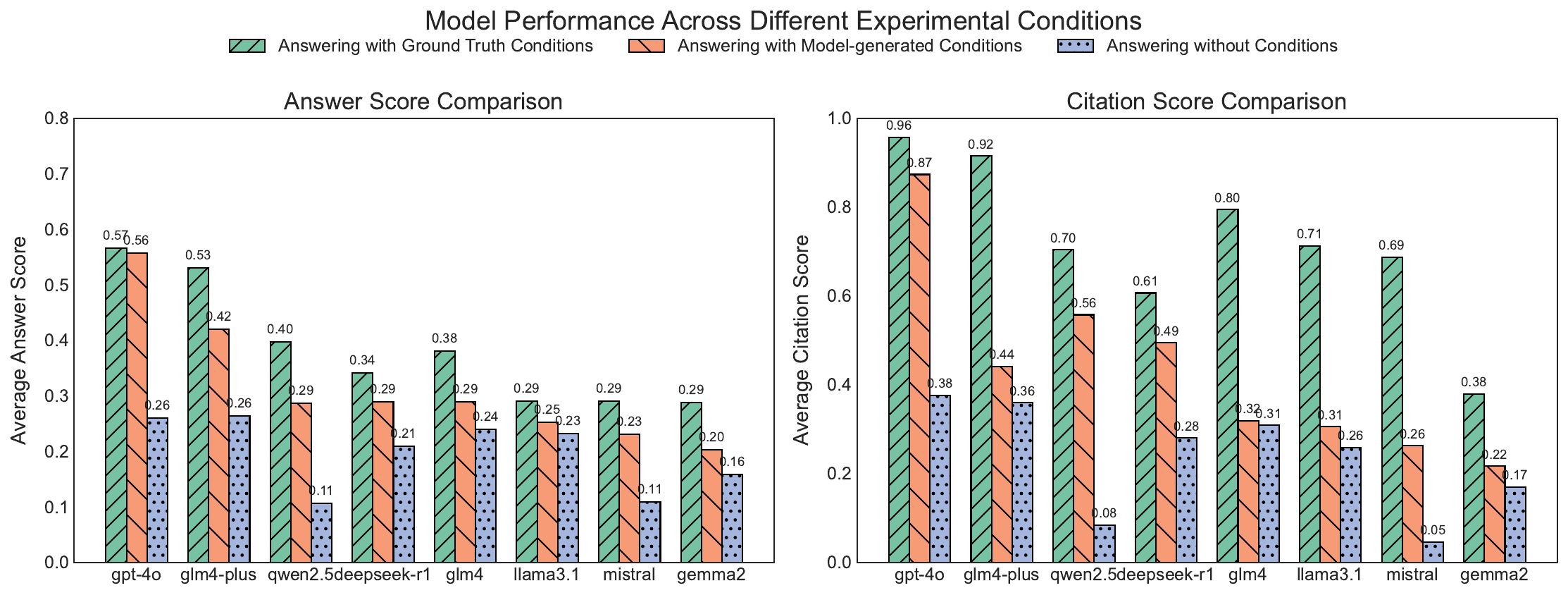}
        \caption{Model performance in Answer Score and Citation Score, comparing answering without conditions, answering based on identified conditions (Main Experiment), and answering based on ground-truth conditions.}
    \label{fig:answer-score-comparison}
\end{figure*}

\subsection{Answer Generation Performance}

Answer generation shows similar variability, with GPT-4o achieving the highest score of $0.558$ ($\sigma = 0.157$), significantly outperforming other models. GLM4-plus follows at $0.420$, and Qwen2.5 leads local models with $0.287$. The performance gradient is steep, with the weakest models (Gemma2 and Mistral) scoring only $0.203$ and $0.231$, respectively. This stark performance gap suggests that proprietary API architectures possess substantially enhanced capabilities for generating accurate answers to ambiguous queries. We further visualise the model performance on four metrics in Figure~\ref{fig:score-bar}, which complements Table~\ref{tab:performance-metrics-separated} by visualising how models vary in their precision vs. coverage trade-offs, especially via Answer Count Difference. 

\subsection{Citation Generation Performance}

Citation generation showed the widest performance gap, revealing GPT-4o's exceptional performance at $0.875$ ($\sigma = 0.207$), followed by Qwen2.5 at $0.558$ ($\sigma = 0.359$) and DeepSeek-R1 at $0.501$ ($\sigma = 0.342$). While API models excel at source attribution, most local models achieve relatively low Citation Scores, with Gemma2 reaching only $0.217$ ($\sigma = 0.277$). This four-fold performance gap suggests local models struggle significantly with accurately attributing information to sources when processing long retrieved passages, while GPT-4o demonstrates a remarkable ability to ground its answers in appropriate citations.

\subsection{Scaling Analysis}

Figure~\ref{fig:expand-distributions} shows the density distributions of the scores, helping to compare the consistency and robustness of the models between metrics. Our findings reveal a clear distinction between proprietary and open-sourced models: API models like GPT-4o show bimodal score distributions, while local models show unimodal distributions with lower variance and lower peak scores. In particular, API models exhibit significantly enhanced capabilities in handling complex queries, with GPT-4o achieving a combined score of $0.662$ and GLM4-plus scoring $0.388$, substantially outperforming the best local model (Qwen2.5 at $0.360$). For condition identification, GPT-4o's scores peak around $0.552$, more than double the average performance of all local models. The score distribution patterns also differ markedly. API models display distinctive bimodal distributions in answer scores, with GPT-4o showing peaks between $0.5$ to $0.7$, whereas local models cluster around $0.2$ to $0.3$. Most notably, GPT-4o shows an unusual spike near $1.0$ in citation scores, indicating perfect citation in many cases, a capability largely absent in local models.

Another interesting pattern emerges in the count differences in answers, shown in Figure~\ref{fig:score-bar}. GPT-4o tends to produce fewer answers than expected ($-0.17$), suggesting a more selective approach, while models like GLM4-plus, GLM4, and Mistral generate significantly more answers ($+1.01$, $+1.08$, and $+1.09$, respectively). This observation may provide clues on models adopting different strategies in handling ambiguity: GPT-4o appears to prioritise precision with fewer, higher-quality answers, while most other models offer broader coverage at the expense of precision.

\begin{figure}[t]
\centering
\includegraphics[width=0.95\columnwidth, trim={0 .5cm 0 1cm}]{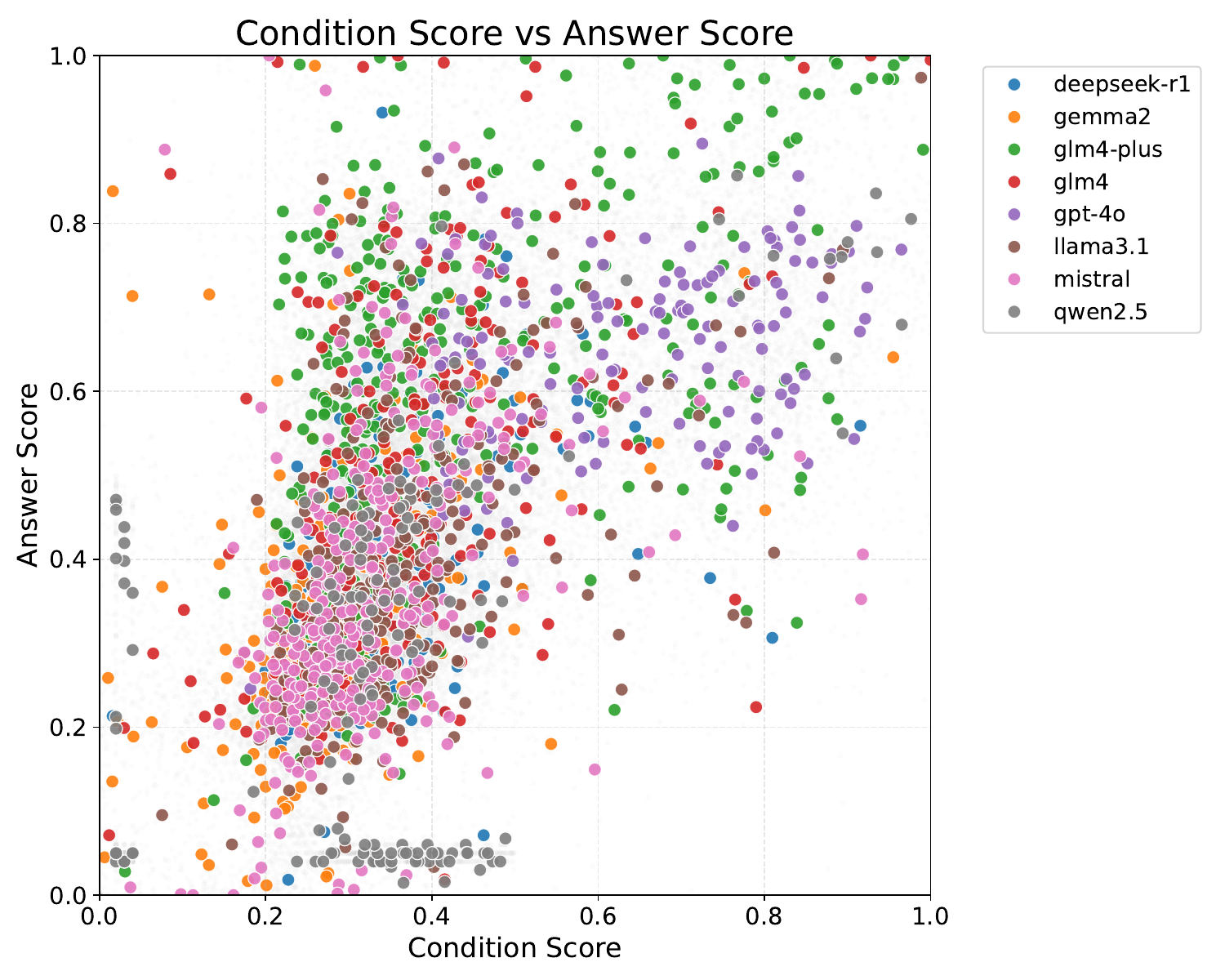}
\caption{Relationship between condition and answer scores across all models. } 
\label{fig:correlation}
\end{figure}

\subsection{Study on the Significance of Conditions}
To validate the importance of conditions in RAG and QA systems, we conducted comparative experiments across three approaches: RAG with self-generated conditions (the same as the main experiment), RAG with annotated ground-truth conditions, and traditional RAG without considering conditions. As shown in Figure~\ref{fig:answer-score-comparison}, both Answer Score and Citation Score demonstrate consistent hierarchical patterns across all tested models. 

In the results, answering with ground-truth conditions consistently yields the highest performance across all models. For answer scores, GPT-4o achieves $0.57$ with ground-truth conditions, compared to $0.56$ with self-generated conditions and $0.26$ without conditions. This pattern holds across all models, with ground-truth conditions providing an average improvement of $0.20$ over the unconditioned baseline. Citation scores show even more drastic improvements, with ground-truth conditions enabling GPT-4o to achieve $0.96$, compared to $0.87$ with self-generated conditions and $0.38$ without conditions, a more than $100\%$ improvement from baseline to optimal conditions.

These results strongly validate our central hypothesis, supported by correlation analysis between condition quality and answer performance (Pearson: $0.598$, Spearman: $0.637$, $p<0.001$). As illustrated in Figure \ref{fig:correlation}, models that achieve higher condition scores consistently demonstrate stronger answer performance, confirming that effective disambiguation through condition identification directly enhances response quality. The inclusion of condition discovery in ambiguous QA, especially with accurate ground-truth conditions, effectively improves both answer quality and citation accuracy. The consistent performance gaps across both metrics underscore the fundamental importance of conditional information in enhancing RAG system performance, with the benefits extending across models of various scales and architectures.

\begin{table*}[htbp]
\centering
\small
\begin{tabular}{lcccc}
\toprule
\textbf{Model} & \textbf{Closed-book} & \makecell{+ \textbf{Model-generated}\\ \textbf{Conditions}} & \makecell{+ \textbf{Ground-truth}\\ \textbf{Conditions}} & \textbf{Improvement} \\
\hline
\multicolumn{5}{l}{\textbf{API Models}} \\
\hline
GPT-4o & 0.25 & 0.56 & 0.57 & $+128\%$ \\
GLM4-Plus & 0.24 & 0.42 & 0.53 &$+121\%$ \\
\hline
\multicolumn{5}{l}{\textbf{Local Models}} \\
\hline
Qwen2.5 (7B) & 0.15 & 0.29 & 0.40 &$+167\%$\\
Mistral (7B) & 0.17 & 0.23 & 0.29 &$+161\%$\\
Gemma2 (9B) & 0.15 & 0.20 & 0.29 &$+93\%$\\
LLaMA3.1 (8B) & 0.14 & 0.25 & 0.29 &$+107\%$\\
GLM4 (9B) & 0.14 & 0.29 & 0.38 &$+171\%$\\
DeepSeek-R1 (7B) & 0.07 & 0.29 & 0.34 &$+400\%$\\
\midrule
Avg. & 0.164 & 0.316 & 0.386 &$+135\%$ \\
\bottomrule
\end{tabular}
\caption{Answer scores on CondAmbigQA of models' losed-book performance vs. condition-grounded reasoning, where no retrieved passages are provided.}
\label{tab:closedbook}
\end{table*}

\subsection{Closed-book Ablation without Retrieval}
To isolate the effect of condition-based reasoning, we evaluate a closed-book setting, where no passages are retrieved or provided as reference; zero-shot direct answering and reasoning with model-hypothesised conditions are tested. The results are reported in Table~\ref{tab:closedbook}. 
It reveals a consistent and substantial drop in answer quality across all models when external context is removed. Models with self-generated conditions (without referencing retrieval) show a $93\%$ increase in the answer score relative to the closed-book baseline, demonstrating that condition reasoning can enhance a model's ability to generate relevant and accurate responses. Compared with results in which ground-truth conditions were provided, we observe even greater gains, with an average improvement of $135\%$ from the closed-book baseline. These results reinforce our hypothesis that many failures in ambiguous QA stem from a lack of contextual grounding rather than inherent deficiencies in the model’s capabilities.

\subsection{Case Study Analysis}
We present a case study section in the Appendix~\ref{appendix_labelf}, which inludes a comprehensive discussion over models' performance patterns and failure cases. Detailed analyses on two ambiguous queries are also provided.

\subsection{Generalisation to External Datasets}
To validate generalisability, we applied our condition-based disambiguation framework using GPT-4o to the ALCE-ASQA dataset (948 questions with Dense Passage Retrieval (DPR)-retrieved passages provided). Despite ALCE-ASQA lacking ground-truth conditions, our method required only minor adaptations. The results demonstrate a clear improvement: direct responses without conditions scored $0.374$, while our condition-based approach achieved $0.471$, a substantial gain of $10\%$. This improvement, combined with the strong correlation between condition quality and answer performance (Pearson: $0.598$, Spearman: $0.637$, $p < 0.001$), confirms that condition-based disambiguation generalises effectively across different ambiguous QA datasets.

\section{Conclusion and Future Work}
This work introduces \textbf{CondAmbigQA}, a novel framework and benchmark designed to address ambiguity in QA by explicitly identifying conditions. Our experiments demonstrate that incorporating explicit condition identification enhances both answer quality and interpretability by clarifying the decision-making process. The analysis reveals that while larger models excel in condition processing, even moderate-sized models gain substantial benefits from this guidance. Additionally, our human-LLM collaborative annotation process has helped ensure a high-quality dataset with reduced subjectivity and bias. Overall, CondAmbigQA establishes a new paradigm for enhancing performance and reliability in ambiguous QA scenarios.

Our findings suggest that condition identification could serve as a foundation for enhancing LLM reasoning capabilities. Future research could integrate condition-based frameworks into the architecture of LLMs to improve their logical reasoning abilities. This could involve developing specialised reasoning mechanisms that focus on condition representations and their logical dependencies.

\section*{Acknowledgement}
This work was supported by Lingnan University, Hong Kong, through the Faculty Research Grant (No. SDS24A2, SDS24A8, SDS24A12, and SDS24A19), Direct Grant (DR25E8), Lam Woo Research Fund (LWP20040), and Shenzhen University-Lingnan University Joint Research Programme 2025/2026 (SZU-LU009/2526), as well as by the Hong Kong Research Grants Council through the Faculty Development Scheme (Project No. UGC/FDS16/E10/23).

\section*{Limitations}
Despite the promising results, several limitations remain:
\begin{itemize}
    \item \textbf{Dataset Representativeness:} While we have expanded our dataset to 2,000 annotated instances through our human-LLM collaborative process, certain types of ambiguity may still be underrepresented. Complex interdependent ambiguities or domain-specific interpretations in specialised fields may require further targeted expansion to ensure comprehensive coverage. Moreover, current annotation process remains resource-intensive and intellectually demanding due to the need for extensive review and cross-checking by experts.
    
    \item \textbf{Performance Gap:} The significant difference between API models (GPT-4o: $0.701$ combined score) and local models (best: Qwen2.5 at $0.469$) indicates that high-quality condition identification may remain challenging for resource-constrained applications. This gap suggests that condition-based disambiguation currently benefits most from advanced model capabilities that may not be widely accessible.
    
    \item \textbf{Generalisation Boundaries:} Although our approach demonstrates effective generalisation to ALCE-ASQA with a $10\%$ improvement, we encountered limitations with datasets lacking passage level references for citation evaluation. The framework may be less effective for inherently subjective or opinion-based queries where multiple interpretations remain equally valid regardless of conditions.
    
    \item \textbf{Real-time Deployment:} The two-stage process of first identifying conditions and then generating answers introduces additional computational overhead that could impact latency in time-sensitive applications. While this approach significantly improves quality, optimising for real-time response in production environments remains challenging.

    \item \textbf{Relatively Heuristic Evaluation Metrics:} Open-domain retrieval may surface conflicting or adversarial passages. Our evaluation penalises unsupported/incoherent generations and rewards condition-separated answers, but explicit adversarial-evidence detection is out of scope. Future work will integrate conflict detection and robustness checks into the pipeline.
\end{itemize}

These limitations highlight the need for future refinement of both the framework and the associated methodologies, ensuring that the benefits of condition-based disambiguation can be maintained across a broader spectrum of applications and model architectures.

\bibliography{anthology}

\appendix

\begin{table*}[t]
\small
\section*{Appendix}
\section{Dataset Examples}  
\label{appendix_label}
\begin{tabular}{p{\textwidth}}
\toprule
\textbf{Question}: When did the show Last Man Standing start? \\
\midrule
\textbf{(1) Condition}: ``Last Man Standing'' is an American sitcom that aired on ABC and Fox. The show originally premiered on ABC in 2011 and was later picked up by Fox in 2018. \\
\textbf{(1) Ground-truth}: The show first premiered on ABC on October 11, 2011, marking its initial broadcast with a special one-hour episode. \\
\textbf{(1) Citations}: \\
Fragment a: ``The show premiered on ABC on October 11, 2011, with a one-hour special episode.'' \\
Fragment b: ``The show originally aired on ABC, then switched to Fox, where it continued in 2018.'' \\
Fragment c: ``Last Man Standing debuted on ABC on October 11, 2011, airing two episodes in the first hour.'' \\[1ex]
\midrule
\textbf{(2) Condition}: ``Last Man Standing'' was canceled by ABC and later re-aired by Fox. The show continued to air after transitioning from ABC to Fox. \\
\textbf{(2) Ground-truth}: On Fox, the show ``started'' again on September 28, 2018, marking its re-premiere. \\
\textbf{(2) Citations}: \\
Fragment x: ``The show's re-premiere occurred on Fox on September 28, 2018.'' \\
Fragment y: ``After being canceled by ABC, Fox picked up the show, with the first new episode airing on September 28, 2018.'' \\
Fragment z: ``Fox aired the first season on September 28, 2018, marking the show's new chapter.'' \\[1ex]
\midrule
\textbf{Retrieval Fragments}: \\
Fragment 1: ``Fox began airing the seventh season on September 28, 2018, after the show's cancellation on ABC.'' \\
Fragment 2: ``The show's first season on Fox premiered on September 28, 2018, following its ABC cancellation.'' \\
Fragment 3: ``Last Man Standing, which had been canceled by ABC, returned for its seventh season on Fox on September 28, 2018.'' \\[1ex]
Fragment 4: ``Last Man Standing debuted on ABC on October 11, 2011, marking its official start.'' \\
Fragment 5: ``The show's premiere on ABC occurred on October 11, 2011, as a one-hour special.'' \\
Fragment 6: ``The show, starring Tim Allen, first aired on ABC in 2011 before transitioning to Fox in 2018.'' \\[1ex]
\bottomrule
\end{tabular}
\caption{An example from our CondAmbigQA dataset.}
\end{table*}

\begin{table*}[t]
\section{Query Prompts Template}
\label{appendix_labelb}
\small
\begin{tabular}{p{\textwidth}}
\toprule
\textbf{Query Analysis Instructions Template} \\
\midrule
You are a professional question analysis assistant. Your task is to analyse questions and their previous incomplete annotations, determining whether these questions contain ambiguities or have multiple possible answers. Please carefully read the following instructions and complete the analysis as required. First, you will receive two inputs:\\[1ex]
\verb|<questions> {{QUESTIONS}} </questions>| \\
\verb|<previous_annotations> {{PREVIOUS_ANNOTATIONS}} </previous_annotations>| \\[1ex]
Please follow these steps: \\[1ex]
1) Read each question and annotation carefully. \\[1ex]
2) Analyse each question for: \\[1ex]
\;\;\;\; a) ambiguity - explain different interpretations \\
\;\;\;\;  b) multiple possible answers - provide examples \\[1ex]
3) Consider: question clarity, vague terms, context sufficiency, subjective elements \\[1ex]
4) Use format: \\[1ex]
\;\;\;\;\verb|<analysis>| \\
\;\;\;\;\verb|<question_number>Number</question_number>| \\
\;\;\;\;\verb|<question_text>Text</question_text>| \\
\;\;\;\;\verb|<ambiguity_analysis>Results</ambiguity_analysis>| \\
\;\;\;\;\verb|<multiple_answers>Results</multiple_answers>| \\
\;\;\;\;\verb|</analysis>| \\[1ex]
5) Compare with previous annotations \\[1ex]
\bottomrule
\end{tabular}
\caption{Instruction template used to analyse queries from ASQA. We use GPT-4o to identify data samples where ambiguity is truly impactful. }
\end{table*}
\clearpage

\begin{table*}[t]
\section{Dataset Prompts}
\label{appendix_labelc}
\centering
\small
\begin{tabular}{p{\textwidth}}
\toprule
\textbf{Dataset Prompts (Part 1)} \\
\midrule
\textbf{Question Answering}: \\
You are tasked with providing a structured answer to a question based on the given text fragments. Your goal is to present possible interpretations supported by the fragments, clearly distinguishing between preconditions and detailed answers. \\[1ex]
Question: <question> [INSERT QUESTION HERE] </question> \\[1ex]
Text fragments: \newline
<fragments> \newline
[INSERT FRAGMENTS HERE] \newline
</fragments> \\[1ex]
Answer format: \\
<answer> \\
Interpretation [X]: \\
Preconditions: \\
* [Necessary background information or assumptions, not directly answering the question] [Fragment X] \\
* [Necessary background information or assumptions, not directly answering the question] [Fragment Y] \\
Detailed answer: \\
* [Specific information directly answering the question] [Fragment Z] \\
* [Specific information directly answering the question] [Fragment A, Fragment B] \\
\mbox{[Repeat the Interpretation structure for as many interpretations as necessary]} \\
</answer> \\[1ex]
Ensure all interpretations are distinct, citing relevant fragments for support. If conflicting information is found, present all viewpoints with sources. \\
\midrule
\textbf{Ambiguity Analysis}: \\
Analyse potential ambiguities in the question ``[INSERT QUESTION HERE]'' based on the provided interpretations. Consider different contexts and how they influence interpretations. \\[1ex]
<analysis> \\
Ambiguity point [X]: [Describe ambiguity that could lead to different interpretations] \\
Impact: \\
1. [Impact on Interpretation 1] [Based on Fragment X, Y] \\
2. [Impact on Interpretation 2] [Based on Fragment Z, A] \\
Contextual considerations: [How different backgrounds might affect understanding] \\
\mbox{[Repeat the Ambiguity point structure for as many ambiguities as necessary]} \\
</analysis> \\[1ex]
Explain how each ambiguity leads to different valid answers, citing relevant fragments. \\
\midrule
\textbf{Evidence Evaluation}: \\
For each interpretation of the question ``[INSERT QUESTION HERE]'', evaluate the supporting evidence. Consider source reliability, consistency across fragments, and potential biases. \\[1ex]
<evaluation> \\
Interpretation [X]: [Brief summary of Interpretation X] \\
Evidence assessment: \\
* Strengths: [List strong evidence supporting this interpretation] [Fragment X, Y] \\
* Weaknesses: [Point out potential issues or shortcomings] [Fragment Z] \\
* Consistency: [Evaluate the consistency of information across fragments] \\
Overall credibility: [Provide an overall assessment, e.g., ``High'', ``Medium'', or ``Low''] \\
\mbox{[Repeat the Interpretation structure for as many interpretations as necessary]} \\
</evaluation> \\[1ex]
Provide a balanced assessment, citing specific fragments to support your evaluation. \\
\bottomrule
\end{tabular}
\caption{The complete sets of dataset-construction prompts provided to annotators (Part 1). GPT-4o is instructed to process each query in the first round of annotation.}
\end{table*}

\begin{table*}[t]
\small
\begin{tabular}{p{\textwidth}}
\toprule
\textbf{Dataset Prompts (Part 2)} \\
\midrule
\textbf{Structured Answer}: \\
Please provide your answer using the following format: \\[1ex]
<answer> \\
Interpretation [X]: \\
Preconditions: \\
* [Necessary background information or assumptions, not directly answering the question] [Fragment X] \\
* [Necessary background information or assumptions, not directly answering the question] [Fragment Y] \\
Detailed answer: \\
* [Specific information directly answering the question] [Fragment Z] \\
* [Specific information directly answering the question] [Fragment A, Fragment B] \\
\mbox{[Repeat the Interpretation structure for as many interpretations as necessary]} \newline
</answer> \\[1ex]

Provide all possible interpretations, ensuring that preconditions and detailed answers are clearly distinct. Every statement must be supported by at least one fragment citation. If you find conflicting information, present all viewpoints and clearly indicate the source of each. \\
\midrule
\textbf{Calibration}: \\
You are tasked with generating a response based strictly on the provided retrieved fragments. Do not introduce any external knowledge or assumptions. Your job is to fill out the following fields using only the information present in the fragments. If any information is missing, leave that field blank. \\[1ex]
1. Condition: Summarise the context of the question strictly using the provided fragments. Do not speculate beyond the given information. \\
2. Ground-truth: Provide the exact answer to the question based on the retrieved fragments. Use only what is explicitly stated. \\
3. Citations: List the relevant fragments that support your answer. Include the title and text of the fragments that were used. \\
4. Reason: Explain how the answer was derived solely from the fragments, and mention why any gaps in information were left unfilled. \\[1ex]
Fragments: {retrieved fragments} \\[1ex]
Output format: \\
{
 ``condition'': ``<summary based on fragments>'',
 ``ground truth'': [``<answer derived from fragments>''],
 ``citations'': [
   {
     ``title'': ``<fragment title>'',
     ``text'': ``<fragment text>''
   }
 ],
 ``reason'': ``<explanation>''
} \\
\midrule
\textbf{Merging}: \\
You are provided with a question and several annotated dictionaries. Your task is to merge all the dictionaries without changing the structure or key names. Consolidate similar information, eliminate redundancy, and ensure that the final output accurately reflects the content of all dictionaries. Do not introduce external knowledge or assumptions. \\[1ex]
Question: {question} \\[1ex]
Dictionaries: {dictionaries} \\[1ex]
Instructions: \\
- Merge the ``condition'' fields from all dictionaries into one, keeping only unique and relevant information. \\
- Merge the ``ground truth'' fields into a single list, ensuring no redundant entries. \\
- Combine the ``citations'' fields from all dictionaries, ensuring all relevant citations are included without duplication. \\
- Leave the ``reason'' field as an empty string. \\[1ex]
Output format: \\
{
 ``condition'': ``<merged condition from all dictionaries>'',
 ``ground truth'': [``<merged ground truth from all dictionaries>''],
 ``citations'': [
   {
     ``title'': ``<citation title from any dictionary>'',
     ``text'': ``<citation text from any dictionary>''
   }
 ],
 ``reason'': 
} \\
\bottomrule
\end{tabular}
\caption{The complete sets of dataset-construction prompts provided to annotators (Part 2). GPT-4o is instructed to process each query in the first round of annotation.}
\end{table*}

\clearpage

\begin{table*}[t]
\vspace{-30mm}
\section{Evaluation Prompts}
\label{appendix_labeld}
\small
\begin{tabular}{p{\textwidth}}
\toprule
\textbf{Evaluation Prompts} \\
\midrule
\textbf{RAG with Conditions Prompt}: \\
Question: \{question\} \\
Retrieved fragments: \\
\{Fragment 1 - \{title\}: \{text\}\} \\
... \\
Please complete the following tasks: \\
1. Identify up to FIVE key conditions related to the question based solely on the provided fragments. \\
2. For each condition, provide a corresponding detailed answer. \\
3. Cite the sources (fragment numbers) that support each condition and answer. \\
4. Output the results in JSON format with the following structure. \\
\midrule
\textbf{Modified Condition-based Prompt}: \\
Question: \{question\} \\
Context fragments: \\
\{Fragment 1 - \{title\}: \{text\}\} \\
... \\
Conditions to address: \\
Condition 1: \{condition\} \\
... \\
IMPORTANT: Respond with ONLY the following JSON format, no other text. \\
\midrule
\textbf{Standard RAG Prompt}: \\
Question: \{question\} \\
Retrieved fragments: \\
\{Fragment 1 - \{title\}: \{text\}\} \\
... \\
Please complete the following tasks: \\
1. Answer the question based solely on the provided fragments. \\
2. Cite up to FIVE sources (fragment numbers) that support your answer. \\
\midrule
\textbf{Evaluation Metrics - Condition Correctness}: \\
- Name: ``Condition Correctness'' \\
- Criteria: ``Determine whether the actual condition is factually correct based on the expected condition.'' \\
- Evaluation steps: \\
1. Check whether the facts in 'actual condition' contradicts any facts in 'expected condition'. \\
2. Heavily penalise omission of critical details in the condition. \\
3. Ensure that the condition is clear and unambiguous. \\
\midrule
\textbf{Evaluation Metrics - Answer Correctness}: \\
- Name: ``Answer Correctness'' \\
- Criteria: ``Determine whether the actual answer is factually correct based on the expected answers.'' \\
- Evaluation steps: \\
1. Check whether the facts in 'actual answer' contradicts any facts in 'expected answers'. \\
2. Heavily penalise omission of critical details in the answer. \\
3. Ensure that the answer directly addresses the question without irrelevant information. \\
\bottomrule
\end{tabular}
\caption{Evaluation prompts. The models are prompted according to these instructions and their outputs are evaluated using the \(\textit{G-Eval}\) function as implemented in the \texttt{DeepEval} package.}
\end{table*}

\clearpage

\section{G-Eval Reliability Analysis}
\label{appendix_labele}

To assess the reliability of G-Eval on our CondAmbigQA benchmark, we conducted a small-scale correlation analysis comparing G-Eval scores against human annotations on 20 randomly sampled examples. Human ratings used the following 10-point rubrics:

\begin{itemize}
  \item \textbf{Condition Quality (1–10)}: how accurately the condition captures ambiguity, covers distinct valid interpretations, and maintains logical coherence.
  \item \textbf{Answer Quality (1–10)}: how accurate, complete under the stated condition, and factually sound (no hallucinations) the answer is.
\end{itemize}

We then calculated Pearson and Spearman correlation coefficients between G-Eval and human scores. The results are presented in Table~\ref{tab:gev_correlation}.

\begin{table}[t]
  \centering
  \small
    \begin{tabular}{lccc}
      \toprule
      \textbf{Metric}   & \textbf{Pearson $\rho$} & \textbf{Spearman $\rho$} & \textbf{p-value} \\
      \midrule
      Condition Quality & 0.88                    & 0.89                     & $<0.001$         \\
      Answer Quality    & 0.83                    & 0.68                     & $<0.01$          \\
      \bottomrule
    \end{tabular}
  \caption{Correlation between G-Eval and human annotations on 20 examples.}
  \label{tab:gev_correlation}
\end{table}

These high correlation coefficients demonstrate that G-Eval closely tracks human judgments in both condition identification and conditional answer quality, validating its use as an automatic evaluator for large-scale ambiguous QA benchmarking.

\section{Case Study Analysis}
\label{appendix_labelf}

Our case studies reveal how different models handle ambiguous queries, with notable variations in performance between API-based models (GPT-4o, GLM4-plus) and local models (LLaMA3.1, Gemma2, GLM4, Qwen2.5). We present detailed analyses of responses to ambiguous questions where multiple valid interpretations exist, focusing on condition identification, answer generation, and citation accuracy.

\subsection{Model Performance on Ambiguous Queries}

We examine model responses to two representative ambiguous queries: ``Which is bigger Kansas City or St. Louis?'' and ``When did color TV come out in US?'' These questions are ambiguous because they can be interpreted in multiple valid ways, requiring models to identify distinct conditions and provide corresponding answers.

For the city comparison query, we identified two key valid interpretations:
\begin{enumerate}
    \item Metropolitan area comparison: Greater St. Louis (2.8 million) is larger than the Kansas City metropolitan area (2.2 million).
    \item City proper comparison: Kansas City has a larger city proper population (approx. 480,000 by 2017) than St. Louis.
\end{enumerate}

For the colour TV question, multiple valid perspectives include:
\begin{enumerate}
    \item Technological introduction: Color TV was officially introduced in December 1953 with the approval of the NTSC standard, with the first national broadcast on January 1, 1954.
    \item Widespread adoption: Color TV became widely adopted in the mid-1960s, with NBC's 1965 transition to colour programming catalysing industry-wide changes.
\end{enumerate}

\subsection{Performance Patterns and Failure Modes}

\begin{table*}[t]
\centering
\small
\begin{tabular}{|p{0.3\textwidth}|p{0.65\textwidth}|}
\hline
\textbf{Model Category} & \textbf{Performance Characteristics} \\
\hline
API Models (GPT-4o, GLM4-plus) & Higher condition quality, better answer accuracy, stronger ability to identify valid interpretations, more precise citations \\
\hline
Local Models (LLaMA3.1, Gemma2, etc.) & Often generate irrelevant conditions, lower answer accuracy, struggle with condition-answer pairs \\
\hline
\end{tabular}
\caption{Key performance differences between model categories.}
\label{tab:model_categories}
\end{table*}

Our analysis reveals distinct patterns of performance as summarised in Table~\ref{tab:model_categories}.
We also identified three key failure patterns across multiple examples.
\begin{enumerate}
    \item \textbf{Condition Misidentification}: Smaller models frequently generate conditions that miss the core ambiguity. For example, Gemma2's response to the city comparison query included ``Influence of both cities in their respective metropolitan areas'' rather than explicitly addressing which city is larger.
    \item \textbf{Factual Inaccuracy}: Models sometimes provide incorrect information. DeepSeek incorrectly stated, ``the Kansas City metropolitan area is larger than Greater St. Louis,'' contradicting available data.
    \item \textbf{Citation Failures}: Most models, particularly local ones, struggle with citation accuracy. Even when answers contain correct information, they often cite wrong fragments, reducing their reliability and trustworthiness.
\end{enumerate}

Using balanced scoring metrics, we established performance thresholds: scores below 0.30 indicate inadequate responses, 0.30 to 0.45 represent partially adequate answers, and above 0.50 indicate high-quality responses.

\subsection{Detailed Analysis: City Comparison Query}

Table~\ref{tab:city_comparison_ground_truth} presents the ground-truth conditions for the city comparison query. Table~\ref{tab:city_comparison_models} shows various model responses to the city comparison query.

\begin{table*}[t]
\centering
\small
\begin{tabular}{|p{0.2\textwidth}|p{0.75\textwidth}|}
\hline
\textbf{Condition} & \textbf{Description} \\
\hline
Metropolitan Comparison & When comparing the metropolitan areas, Greater St. Louis is larger than the Kansas City metropolitan area. Greater St. Louis is the largest metropolitan area in Missouri, with a population of over 2.8 million people. The Kansas City metropolitan area is the second-largest, with a population of more than 2.2 million people. \\
\hline
City Proper Comparison & When comparing the city proper populations, Kansas City, Missouri, is larger than St. Louis, Missouri. Kansas City has a city proper population that has grown to almost 480,000 people by 2017, reflecting steady growth over the years. In contrast, St. Louis has a smaller city proper population. \\
\hline
\end{tabular}
\caption{Ground-truth conditions for the city comparison query}
\label{tab:city_comparison_ground_truth}
\end{table*}

\begin{table*}[t]
\centering
\small
\begin{tabular}{|p{0.1\textwidth}|p{0.2\textwidth}|p{0.35\textwidth}|c|p{0.15\textwidth}|}
\hline
\textbf{Model} & \textbf{Generated Condition} & \textbf{Answer Excerpt} & \textbf{Score} & \textbf{Analysis} \\
\hline
GPT-4o & Comparison of Metropolitan Areas & ``When comparing the size of Kansas City and St. Louis, it is important to consider the metropolitan areas... The Greater St. Louis area is generally considered larger...'' & $0.69$ & Correctly identifies metropolitan comparison \\
\hline
GPT-4o & Comparison of City Proper Populations & ``Kansas City has a larger population...'' & $0.64$ & Accurately distinguishes city proper \\
\hline
GLM4-plus & Comparison of Metropolitan Areas & ``The Greater St. Louis metropolitan area is a bi-state region... St. Louis is the focus of the largest metro area in Missouri...'' & $0.70$ & Thorough comparison with citations \\
\hline
GLM4-plus & Comparison of City Proper Populations & ``Kansas City's city proper population had reached almost 480,000 residents...'' & $0.59$ & Correctly addresses city populations \\
\hline
Gemma2 & Population size comparison & ``St. Louis is indicated to be larger than Kansas City, Missouri...'' & $0.42$ & Confuses historical and current size \\
\hline
Gemma2 & Influence of cities & ``Both Kansas City and St. Louis are anchors for large metropolitan areas...'' & $0.30$ & Doesn't address size comparison \\
\hline
LLaMA3.1 & Kansas City metropolitan area population & ``The Kansas City metropolitan area's population is expected to grow from 2.1 Million to over 2.7 Million by 2040...'' & $0.33$ & Incorrect metropolitan size conclusion \\
\hline
LLaMA3.1 & Greater St. Louis location & ``According to Fragment 1, Greater St. Louis is a bi-state metropolitan statistical area...'' & $0.21$ & Fails to address size comparison \\
\hline
DeepSeek & Population Comparison & ``Based on historical data, the Kansas City metropolitan area is larger than Greater St. Louis...'' & $0.31$ & Incorrect metropolitan comparison \\
\hline
DeepSeek & Historical Growth & ``St. Louis experienced significant population growth in the mid-19th century...'' & $0.29$ & Discusses irrelevant historical context \\
\hline
\end{tabular}
\caption{Model-generated conditions and evaluation for city comparison query}
\label{tab:city_comparison_models}
\end{table*}

\subsection{Detailed Analysis: Color TV Query}

Table~\ref{tab:color_tv_ground_truth} presents the ground-truth conditions for the colour TV introduction query. Table~\ref{tab:color_tv_models} shows various model responses to the colour TV query.

\begin{table*}[t]
\centering
\small
\begin{tabular}{|p{0.15\textwidth}|p{0.8\textwidth}|}
\hline
\textbf{Condition} & \textbf{Description} \\
\hline
Technological Introduction & Color television was officially introduced in the US with the approval of the NTSC standard in December 1953. This allowed for the first national color broadcast on January 1, 1954, featuring NBC's coverage of the Tournament of Roses Parade. Despite this technological milestone, the high cost of color television sets and limited programming meant that consumer adoption was slow. \\
\hline
Widespread Adoption & Color television became widely adopted in the US during the mid-1960s. The transition to color programming gained momentum in 1965 when NBC announced that its prime-time schedule would be almost entirely in color. This prompted other networks to follow suit, leading to a significant increase in color broadcasts. By 1972, more than half of all U.S. households owned a color television. \\
\hline
\end{tabular}
\caption{Ground-truth conditions for the colour TV query}
\label{tab:color_tv_ground_truth}
\end{table*}

\begin{table*}[htbp]
\centering
\small
\begin{tabular}{|p{0.1\textwidth}|p{0.15\textwidth}|p{0.4\textwidth}|c|p{0.15\textwidth}|}
\hline
\textbf{Model} & \textbf{Generated Condition} & \textbf{Answer Excerpt} & \textbf{Score} & \textbf{Analysis} \\
\hline
GPT-4o & Introduction of Color Television & ``Color television was introduced in the US following significant regulatory and technological developments... The NTSC color standard, approved in 1953... The first broadcast was on January 1, 1954...'' & $0.55$ & Provides accurate timeline and key events \\
\hline
GPT-4o & Consumer Adoption and Market Penetration & ``While the technology was introduced in 1953, its widespread adoption was slow... It wasn't until the mid-1960s that color television saw significant consumer adoption...'' & $0.45$ & Correctly addresses adoption timeline \\
\hline
GLM4-plus & Introduction of color television & ``Color television was officially introduced in the United States in 1953 when the FCC approved the NTSC color standard. The first national color broadcast occurred on January 1, 1954...'' & $0.65$ & Clear, accurate introduction account \\
\hline
GLM4-plus & Widespread adoption of color TV & ``Widespread adoption took longer despite its introduction in 1953. It was not until the mid-1960s that color sets started selling in large numbers...'' & $0.63$ & Thorough explanation of adoption timeline \\
\hline
Gemma2 & When were color television broadcasts introduced & ``The first national color broadcast in the US occurred on January 1, 1954. While limited programming was available soon after, it wasn't until the early 1970s that color television widely outsold black-and-white sets.'' & $0.44$ & Contains accurate facts but lacks regulatory context \\
\hline
Gemma2 & Initial factors hindering adoption & ``High prices for color television sets and a scarcity of color programming significantly slowed the acceptance of color television...'' & $0.34$ & Addresses adoption barriers but not the timeline \\
\hline
LLaMA3.1 & Color television sets were initially expensive & ``The high prices of color television sets, combined with the scarcity of color programming, greatly slowed their acceptance in the marketplace...'' & $0.43$ & Addresses barriers to adoption \\
\hline
LLaMA3.1 & First national color broadcast & ``The first national color broadcast was the 1954 Tournament of Roses Parade, which took place on January 1, 1954...'' & $0.38$ & Provides broadcast date but limited context \\
\hline
DeepSeek & When were color TVs first made available & ``Color television sets became available for sale starting in mid-1950s, with the first all-color prime-time season beginning in 1966.'' & $0.29$ & Imprecise timeline and limited details \\
\hline
DeepSeek & When did the first national color broadcast occur & ``The first national color broadcast occurred on January 1, 1954, with NBC transmitting the Tournament of Roses Parade.'' & $0.44$ & Accurate broadcast date but lacks context \\
\hline
\end{tabular}
\caption{Model-generated conditions and evaluation for colour TV query}
\label{tab:color_tv_models}
\end{table*}

\subsection{Comparing DeepSeek Reasoning with Base Models}

An important dimension of our analysis is the comparison between DeepSeek's reasoning-enhanced model and other base models. DeepSeek represents an attempt to improve reasoning capabilities in LLMs through specialized training and architectural modifications. Our case studies reveal significant differences in performance, as shown in Table~\ref{tab:deepseek_comparison}.

The DeepSeek reasoning model demonstrates some improvements over other local models, particularly in its attempt to structure responses more systematically. When addressing the color TV question, DeepSeek formulated conditions as direct questions: ``When were color TVs first made available to the public in the U.S.?'' and ``When did the first national color broadcast occur in the U.S.?'' This approach shows a clearer understanding of the task structure.

However, DeepSeek still falls significantly short of API models in three critical areas:

\begin{enumerate}
    \item \textbf{Factual accuracy}: DeepSeek incorrectly claimed that ``the Kansas City metropolitan area is larger than Greater St. Louis,'' contradicting established facts.
    
    \item \textbf{Condition comprehensiveness}: DeepSeek failed to adequately address both interpretations of the city comparison question, focusing on superficial aspects like ``Historical Growth'' rather than comprehensive size comparisons.
    
    \item \textbf{Answer depth}: While DeepSeek provided some accurate information (e.g., the date of the first color broadcast), its answers lacked the contextual depth and nuance found in API model responses.
\end{enumerate}

These findings suggest that while specialized reasoning training provides some benefits, it does not close the substantial capability gap between local models and larger API models for condition-based RAG tasks.

\subsection{Key Findings}

Our case studies demonstrate significant performance gaps between model categories in condition-based RAG:

\begin{itemize}
    \item \textbf{API models (GPT-4o, GLM4-plus)} consistently identify the core ambiguities in questions and generate conditions that address multiple valid interpretations. Their answer quality is substantially higher, with scores frequently above $0.60$.
    
    \item \textbf{DeepSeek Reasoning model} shows some structural improvements over other local models but still struggles with factual accuracy and comprehensive condition identification. Its performance scores (typically $0.29$-$0.44$) position it marginally better than other local models but far below API models.
    
    \item \textbf{Other local models} often miss key ambiguities, providing either irrelevant conditions or incorrect answers. Their condition and answer quality scores typically fall between $0.21$-$0.45$, indicating partial adequacy at best.
    
    \item \textbf{Citation accuracy} varies dramatically, with API models more likely to correctly cite supporting evidence (50-100\% accuracy), while local models including DeepSeek frequently cite inappropriate or irrelevant fragments ($0$-$50\%$ accuracy).
\end{itemize}

These findings highlight the critical importance of model capability in condition-based RAG systems. When dealing with ambiguous queries, larger API models demonstrate significantly greater ability to identify valid interpretations, generate appropriate conditions, provide accurate answers, and cite relevant evidence. While reasoning-enhanced models like DeepSeek show incremental improvements, the capability gap remains substantial, suggesting that deploying high-capability models is essential for effective condition-based RAG systems, particularly for domains where query ambiguity is common.

\begin{table*}[t]
\centering
\small
\begin{tabular}{|p{0.2\textwidth}|p{0.25\textwidth}|p{0.2\textwidth}|p{0.25\textwidth}|}
\hline
\textbf{Aspect} & \textbf{DeepSeek Reasoning} & \textbf{Other Local Models} & \textbf{API Models} \\
\hline
Condition Identification & Attempts to identify meaningful conditions but often misses key ambiguities (score: $0.29$-$0.32$) & Generate overly generic or tangential conditions (score: $0.21$-$0.33$) & Successfully identify critical ambiguities (score: $0.55$-$0.82$) \\
\hline
Answer Accuracy & Provides accurate details in some cases but often draws incorrect conclusions (score: $0.22$-$0.44$) & Frequently mixes correct and incorrect information (score: $0.24$-$0.45$) & Consistently provides accurate answers (score: $0.44$-$0.78$) \\
\hline
Citation Precision & Low to moderate ($25$-$50\%$) & Very low ($0$-$30\%$) & Moderate to high ($25$-$100\%$) \\
\hline
\end{tabular}
\caption{Comparison of DeepSeek Reasoning with other model categories}
\label{tab:deepseek_comparison}
\end{table*}

\end{document}